\documentclass[letterpaper]{article}
\usepackage[preprint]{aaai2027}
\usepackage[hyphens]{url}
\usepackage{graphicx}
\usepackage{booktabs}
\usepackage{amsmath}
\usepackage{amssymb}
\usepackage{multirow}
\usepackage{array}
\usepackage{placeins}
\usepackage{natbib}
\usepackage{float}
\title{PLRS-IC: A Dual-Calibration Framework for Chest X-Ray Vision-Language Alignment}

\author{
Qixing Zhao,
Jinpeng Li\textsuperscript{*}
}

\affiliations{
South China University of Technology\\
lijinpeng@scut.edu.cn
}

\begin{document}
\maketitle

\begin{abstract}
Fine-grained vision-language alignment in chest radiography enables zero-shot classification, grounding, and segmentation without task-specific annotations. However, this alignment is fundamentally hindered by two intertwined sources of ambiguity: projection-induced visual mismatch and patient-agnostic semantic overlap. First, at the \emph{local} feature level, frontal and lateral radiographs exhibit distinct appearances for the same clinical finding, rendering a shared patch-text similarity geometry inherently suboptimal. Compounding this visual ambiguity is a semantic mismatch during \emph{global} contrastive optimization, where instance-level objectives penalize cross-patient pairs as strict negatives even when they share identical positive clinical concepts. To address this dual ambiguity, we propose PLRS-IC, a unified dual-calibration framework for chest X-ray representation learning. At the local alignment stage, Projection-Conditioned Low-Rank Residual Similarity (PLRS) dynamically adapts patch-text matching to projection-specific manifolds using a bounded, parameter-efficient low-rank residual. At the global optimization stage, Information-Content-Calibrated Soft False-Negative Suppression (IC-SFNS) leverages a corpus-derived information-theoretic prior to soften the penalty of semantically overlapping negatives without altering original contrastive assignments. Extensive experiments across nine public zero-shot benchmark settings demonstrate that our framework yields consistent improvements in classification, grounding, and segmentation, validating the necessity of dual-calibration in medical vision-language pre-training.
\end{abstract}

\section{Introduction}

Chest radiography is routinely used to assess a broad range of thoracic conditions. Obtaining detailed chest X-ray annotations is costly because the process requires clinical expertise. The burden increases further when annotations must also specify the locations of clinical findings. Vision-language pre-training offers an alternative by learning from paired radiographs and reports \cite{convirt,refers}. It supports zero-shot medical image understanding without requiring a separate classifier for each finding \cite{chexzero,gloria}. Zero-shot chest X-ray understanding therefore depends on aligning free-text clinical queries with the visual evidence that supports them.

\begin{figure}[t]
    \centering
    \includegraphics[width=\columnwidth]{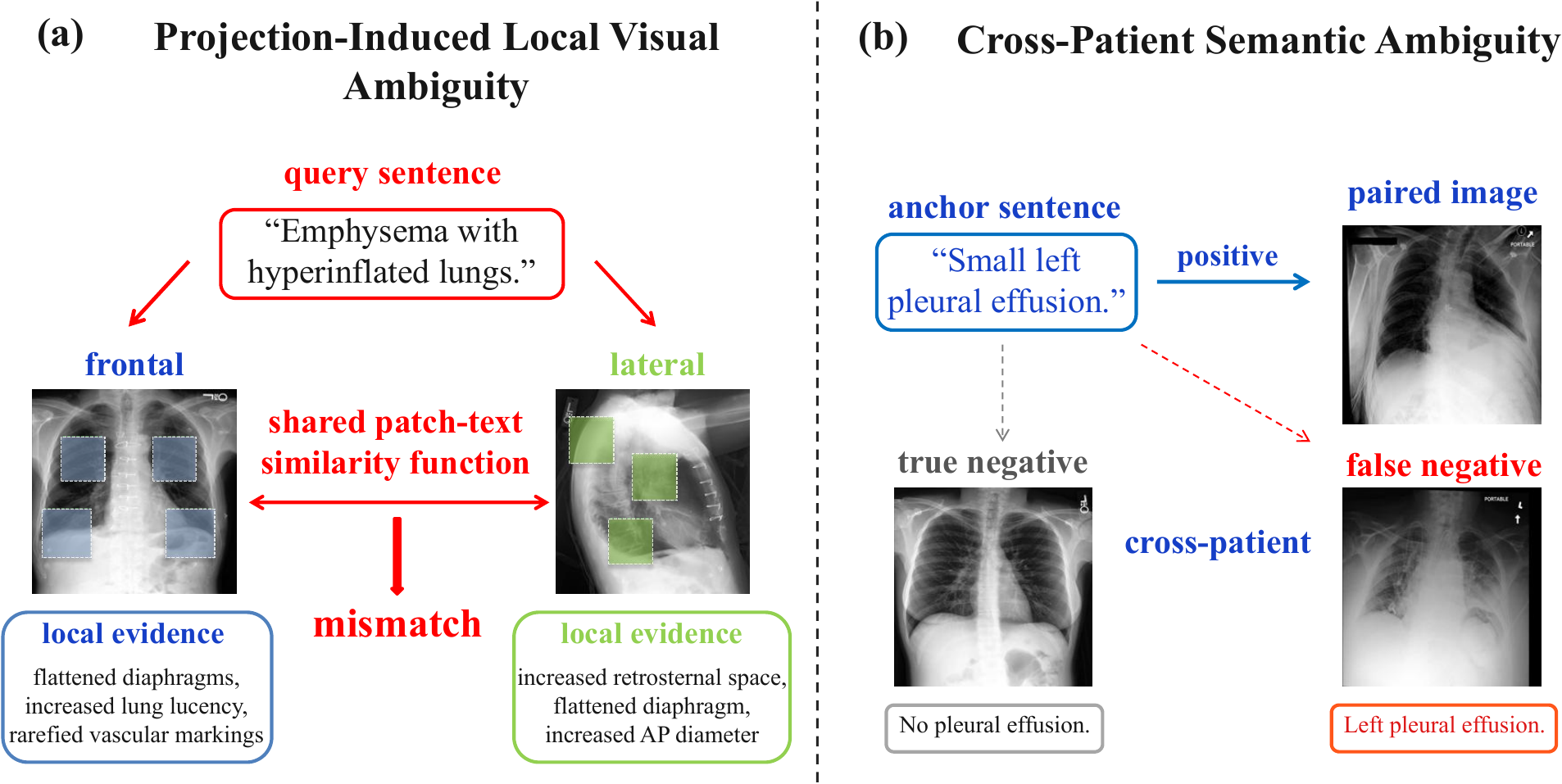}
    \caption{
    \textbf{The dual-ambiguity in chest X-ray vision-language alignment. (a) Local ambiguity.} Frontal and lateral radiographs are independently compared with the same query using a shared patch-text similarity function, despite projection-specific local evidence. \textbf{(b) Global ambiguity.} CLIP-style contrastive learning treats cross-patient pairs as negatives, creating a potential false negative when another patient shares the same positive finding.
    }
    \label{fig:motivation}
\end{figure}

Fine-grained image-text alignment is essential because a finding sentence often refers to evidence confined to a local region \cite{maco,adamatch}. Recent medical vision-language methods extend global image-report matching through phrase-level or patch-level interactions \cite{gloria,mgca,radzero}. Yet, this fine-grained alignment is hindered by projection-induced visual ambiguity. Chest radiographs are acquired under different projections, and the same finding can produce distinct local evidence across views. As shown in Figure~\ref{fig:motivation}(a), frontal manifestations of emphysema include increased lung lucency and rarefied vascular markings. The lateral view instead emphasizes increased retrosternal space and an enlarged anteroposterior diameter. Despite these differences, existing methods compare both projections with the same query through a shared patch-text similarity function. A shared similarity geometry struggles to accommodate projection-dependent appearances across both views, which inevitably reduces the precision of phrase-region correspondence. \emph{This necessitates explicit projection conditioning at the local similarity level to dynamically calibrate visual matching.}

Compounding this local visual ambiguity is a patient-agnostic semantic mismatch introduced during global contrastive optimization. Standard CLIP-style objectives treat a finding sentence and its source image as a positive pair, while images from other patients serve as negatives. As shown in Figure~\ref{fig:motivation}(b), the anchor sentence ``Small left pleural effusion'' is correctly paired with its source image. Another patient's image can depict the identical finding but still remain a strict negative, alongside a true negative without pleural effusion. Treating these cross-patient pairs as equally reliable negatives introduces false-negative supervision and pushes clinically consistent image-text evidence apart \cite{conns,fane}. While existing methods address this problem by changing pair roles or mining latent positives \cite{conns,fane}, sharing a positive concept does not imply complete clinical or visual equivalence between patients. \emph{We instead investigate whether unreliable cross-patient negatives can be softly calibrated using an information-theoretic prior, preserving their original assignments while mitigating semantic overlap.}

To address this dual ambiguity, we introduce PLRS-IC, a unified dual-calibration framework for zero-shot chest X-ray vision-language alignment. PLRS dynamically conditions local patch-text similarity on the radiographic projection, rectifying the visual mismatch. Simultaneously, IC-SFNS calibrates selected cross-patient negative terms using corpus-derived concept statistics, softening the backward semantic mismatch. The two components operate in tandem on local similarity estimation and global contrastive supervision. Our main contributions include:
\begin{itemize}
    \item We introduce PLRS, a projection-conditioned low-rank correction for patch-text similarity. It adapts local matching to projection-specific visual manifolds with few additional parameters, while preserving the base similarity and bounding the correction magnitude.
    
    \item We propose IC-SFNS, an optimization strategy to downweight selected cross-patient negatives. The attenuation weights are derived from a corpus-level concept information-content prior, achieving soft false-negative suppression while the original pair labels remain.
    
    \item We evaluate PLRS-IC on nine public zero-shot chest X-ray benchmarks for classification, grounding, and segmentation, demonstrating consistent performance improvements and validating the efficacy of our dual-calibration framework.
\end{itemize}

\section{Related Work}

\paragraph{Fine-Grained Chest X-Ray Vision-Language Alignment.}
Chest X-ray vision-language pre-training learns from paired
radiographs and reports without dense task-specific annotations
\cite{refers,mrm,biovil,g2d}. GLoRIA and MGCA extend global alignment through region-word or multi-granularity supervision \cite{gloria,mgca}, while CARZero and RadZero further model local image-text interactions for zero-shot interpretation \cite{carzero,radzero}. RadZero directly derives similarity maps for classification, grounding, and segmentation. These methods improve local feature interaction and aggregation, but a shared patch-text similarity function leaves projection-induced local ambiguity unresolved. PLRS addresses this local ambiguity by conditioning the similarity geometry on the known projection of each radiograph.

\paragraph{Projection-Aware Modeling of Chest Radiographs.}
Frontal and lateral radiographs provide complementary anatomical evidence and have motivated multi-view chest X-ray vision-language models. CXR-CLIP uses multiple images and report sections from the same radiographic study to construct study-level supervision \cite{cxrclip}. This strategy expands the available image-text combinations, but does not explicitly distinguish projection-specific local appearances. Med-ST introduces a Mixture of View Experts to encode frontal and lateral radiographs \cite{medst}. It combines the view-specific features through cross-view integration and global-local image-text alignment. Although the experts retain projection-related information, the architecture fuses radiographs that are jointly available from the same study. Pathology-relevant patches have also been aggregated across multiple radiographic views \cite{qiao2026multiview}. Their Frontal-Lateral Alignment preserves view-specific pathological features while encouraging semantic consistency across projections. These approaches integrate complementary evidence from paired views, whereas our setting concerns projection-induced ambiguity within the local alignment of a single radiograph. PLRS calibrates this local similarity according to the known projection without requiring paired frontal and lateral inputs.

\begin{figure*}[!t]
    \centering
    \includegraphics[width=\textwidth]{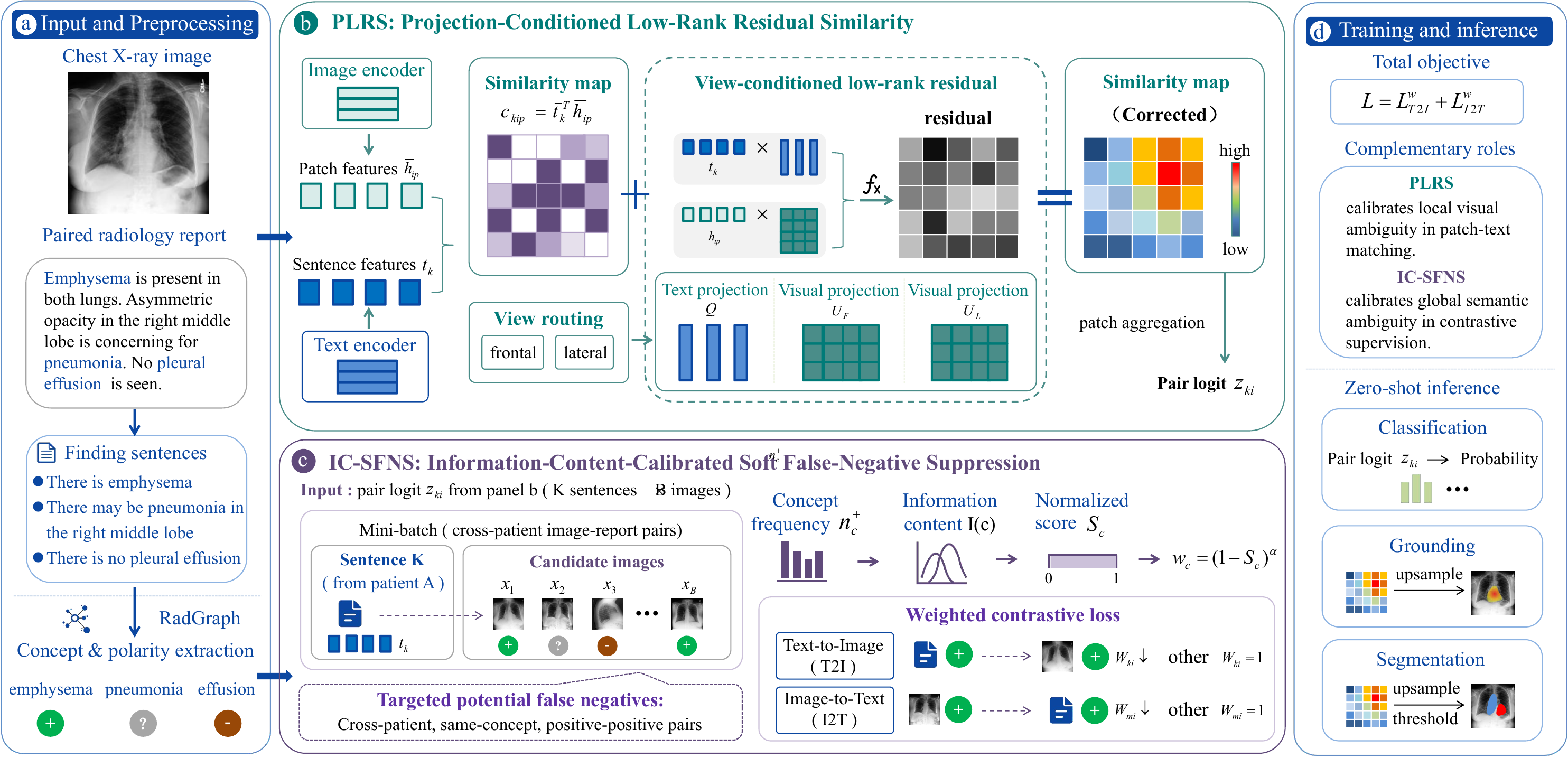}
    \caption{\textbf{Overview of the PLRS-IC framework.} PLRS addresses projection-induced local ambiguity by calibrating patch-text similarity, while IC-SFNS mitigates cross-patient semantic ambiguity by calibrating selected negative contributions during global contrastive optimization.}
    \label{fig:method}
\end{figure*}

\paragraph{False-Negative Handling in Medical Contrastive Learning.}
Medical image-text contrastive learning usually treats paired samples as positives and cross-patient samples as negatives \cite{clip,lit}. This assumption becomes unreliable when different patients share the same positive clinical finding. MedCLIP relaxes strict pair-based supervision by constructing semantic matching targets from clinical labels \cite{medclip}. These targets broadly redefine image-text relatedness instead of selectively adjusting unreliable cross-patient negative terms. Recent methods incorporate richer clinical semantics into contrastive supervision \cite{clipclinic}. CoNNS constructs a hierarchical concept ontology and assigns cross-patient relationships according to their clinical semantics \cite{conns}. FaNe identifies latent positives through report-level semantic similarity and incorporates them into a multi-positive contrastive objective \cite{fane}. Both methods reduce false-negative noise by changing the relation or training role of selected image-text pairs. We instead view cross-patient concept overlap as a source of global semantic ambiguity in contrastive supervision. Because sharing a positive concept is not sufficient to establish complete clinical or visual equivalence, IC-SFNS preserves the original pair assignment and calibrates only its negative contribution. The calibration strength is determined by corpus-derived concept information content.

\section{Method}

Figure~\ref{fig:method} illustrates the PLRS-IC workflow. Given paired radiographs and finding sentences, PLRS produces projection-conditioned patch-text similarity maps and image-sentence logits. IC-SFNS then calibrates selected cross-patient negative terms when these logits enter the bidirectional contrastive objective. The corrected maps and logits support zero-shot classification, grounding, and segmentation.

\subsection{Local Calibration with Projection-Conditioned Low-Rank Residual Similarity}

\paragraph{Base local alignment.}
Consider a minibatch of radiographs $\mathcal{X}=\{x_i\}_{i=1}^{B}$ and finding sentences $\mathcal{Q}=\{q_k\}_{k=1}^{K}$. Because one radiograph may correspond to multiple sentences, $g(k)$ denotes the image paired with sentence $q_k$. The image encoder extracts patch features $\mathbf{h}_{ip}\in\mathbb{R}^{D}$, where $p\in\{1,\ldots,P\}$ indexes the image patches. The text encoder produces a sentence representation $\mathbf{t}_k\in\mathbb{R}^{D}$. The base patch-text similarity is

\begin{equation}
c_{kip}
=
\bar{\mathbf t}_k^{\top}
\bar{\mathbf h}_{ip},
\label{eq:base}
\end{equation}
where $\bar{\mathbf t}_k$ and $\bar{\mathbf h}_{ip}$
denote the $\ell_2$-normalized sentence and patch
representations, respectively.
The patch-level similarity map is aggregated into the image-sentence logit $z_{ki}$. 

\paragraph{Projection-conditioned residual.}
Projection-induced local ambiguity arises because frontal and lateral radiographs express the same clinical finding through different local visual patterns. Although the image encoder may represent part of this variation, Eq.~\eqref{eq:base} evaluates both projections within the same cosine-similarity geometry. PLRS addresses this local ambiguity by conditioning patch-text matching on the known radiographic projection.

Let $v_i\in\{F,L\}$ denote a reliable projection label for
image $x_i$, where $F$ includes frontal projections such as
posteroanterior and anteroposterior views, and $L$ denotes
lateral views.
PLRS introduces a shared text projection
$\mathbf Q\in\mathbb{R}^{D\times d_r}$ and two
view-specific visual projections
$\mathbf U_F,\mathbf U_L\in\mathbb{R}^{D\times d_r}$,
where $d_r\ll D$ is the residual rank.
We use $\mathbf U_{v_i}=\mathbf U_F$ for frontal images and
$\mathbf U_{v_i}=\mathbf U_L$ for lateral images.
When a reliable frontal or lateral label is unavailable, the
routed visual projection is set to zero, so that the residual
branch is bypassed and the base cosine similarity is retained.
The shared projection $\mathbf Q$ preserves a common textual
space, while $\mathbf U_{v_i}$ adapts visual matching to the
observed projection. The low-rank residual score is

\begin{equation}
r_{kip}
=
\frac{
\left(
\mathbf Q^{\top}\bar{\mathbf t}_k
\right)^{\top}
\left(
\mathbf U_{v_i}^{\top}\bar{\mathbf h}_{ip}
\right)
}{
\sqrt{d_r}
}.
\label{eq:residual}
\end{equation}

This formulation represents a projection-conditioned low-rank bilinear interaction. It models additional text-visual correspondence without replacing the base cosine similarity. The factor $\sqrt{d_r}$ stabilizes the residual scale across different rank settings.

\paragraph{Bounded similarity correction.}
PLRS adds the residual to the base similarity as follows:

\begin{equation}
\widetilde c_{kip}
=
c_{kip}
+
\gamma\tanh(r_{kip}),
\label{eq:corrected}
\end{equation}

where $\gamma>0$ is a fixed residual scale. Since $\tanh(r_{kip})\in[-1,1]$, the correction magnitude is bounded by $\gamma$. This bound prevents the residual branch from dominating the base cosine similarity.

The visual projection matrices $\mathbf U_F$ and $\mathbf U_L$ are initialized to zero. This initialization gives $r_{kip}=0$ and $\widetilde c_{kip}=c_{kip}$ at the start of training. PLRS thus begins with the base alignment function and gradually learns projection-specific corrections. The text projection $\mathbf Q$ is initialized from a normal distribution. The zero-initialized visual branches preserve the initial identity mapping.

\paragraph{Integration and parameter efficiency.}
The corrected similarity map
$\widetilde{\mathbf C}_{ki}
=
\{\widetilde c_{kip}\}_{p=1}^{P}$
replaces the original patch-text similarity map in the unchanged aggregation module. The module aggregates this map into the image-sentence logit $z_{ki}$ for contrastive learning. The corrected map is also retained for zero-shot grounding and segmentation.

PLRS introduces $3Dd_r$ trainable projection parameters through $Q$, $U_F$, and $U_L$. The residual scale $\gamma$ is a fixed scalar hyperparameter. Because $d_r$ is much smaller than the feature dimension $D$, the additional trainable parameter cost remains limited. PLRS thus provides the local calibration branch of PLRS-IC, adapting projection-dependent similarity without duplicating the image encoder or requiring paired frontal and lateral inputs.

\subsection{Global Calibration with Soft False-Negative Suppression}

\paragraph{Concept-polarity representation.}
Cross-patient semantic ambiguity arises when images from
different patients share the anchor sentence's positive concept
but remain negatives in the contrastive batch.
IC-SFNS identifies these cases through clinical concept-polarity
overlap and calibrates their negative contributions without
changing the original pair assignments.

RadGraph is applied offline to extract clinical entities and
polarity states from each finding sentence.
After normalization, the entities are mapped to a fixed inventory
$\mathcal C$ containing 25 canonical chest X-ray concepts.
Each successfully mapped sentence $q_k$ is represented by
$(c_k,\pi_k)$, where $c_k\in\mathcal C$ and
$\pi_k\in\{+,-,?\}$ denotes positive, negative, or uncertain
polarity.
Sentences that cannot be mapped to the inventory are not
selected for suppression.

Sentence-level annotations are further aggregated into an
image-level concept-polarity state
$y_i(c)\in\{+,-,?,\varnothing\}$.
The symbol $\varnothing$ indicates that the source report
does not identify concept $c$.
Images associated with the same report inherit the same
annotations, which are used only for training-time negative
calibration.

\paragraph{Information-content prior.}
Not all cross-patient concept overlaps create the same degree of global semantic ambiguity. Common findings may occur in many patients with otherwise unrelated clinical profiles, whereas a rare positive finding provides more specific evidence of semantic overlap. We therefore use corpus-level information content to calibrate the strength of global negative attenuation.

For each concept $c\in\mathcal C$, let $n_c^+$ denote its number of positive occurrences in the training corpus, and let $N$ denote the total number of training samples used to construct the prior. The empirical positive occurrence rate is

\begin{equation}
p(c)=\frac{n_c^+}{N}.
\end{equation}

A larger $p(c)$ indicates that concept $c$ occurs positively in a larger proportion of the training corpus. The information content of concept $c$ is

\begin{equation}
I(c)
=
-\log p(c).
\label{eq:ic}
\end{equation}

Frequent concepts have lower information content, whereas rare concepts have higher information content. We normalize these values across the fixed concept inventory:

\begin{equation}
S_c
=
\frac{
I(c)
}{
\max_{c'\in\mathcal C}I(c')
}.
\label{eq:normalized_ic}
\end{equation}

The normalized score satisfies $0\leq S_c\leq 1$. A score closer to one represents a rarer and more informative concept. The concept-dependent attenuation weight is

\begin{equation}
w_c
=
(1-S_c)^{\alpha}.
\label{eq:weight}
\end{equation}

The exponent $\alpha$ controls the strength of concept-dependent attenuation. A common concept has a relatively small $S_c$, so its weight $w_c$ remains close to one. Its negative contribution is therefore attenuated only weakly. A rare concept has a larger $S_c$ and receives a smaller $w_c$. Its negative contribution is attenuated more strongly when two patients share the same positive concept.

The information-content score determines only the attenuation magnitude. It is not a calibrated probability that an image-text pair is a false negative. All concept statistics are computed once from the training corpus and remain fixed throughout optimization.

\paragraph{Targeted pair weighting.} Let $a_i$ denote the patient identity associated with image $x_i$. For anchor sentence $q_k$ and candidate image $x_i$, IC-SFNS defines

\begin{equation}
W_{ki}
=
\begin{cases}
w_{c_k},
&
\substack{
a_i\neq a_{g(k)},\;
\pi_k=+,\\
y_i(c_k)=+
},
\\[2mm]
1,
&
\text{otherwise}.
\end{cases}
\label{eq:pairweight}
\end{equation}

Attenuation first requires the anchor sentence and candidate image to come from different patients. In addition, both reports must identify the anchor concept with positive polarity. All other image-text pairs retain unit weight.

We restrict attenuation to positive-positive concept collisions because positive mentions indicate that the finding is present in both patients. The two images may therefore contain clinically consistent visual evidence for the same concept. A negative-negative match only indicates that both reports deny the finding. Absence does not define a shared local pattern and can occur in otherwise unrelated radiographs. Attenuating these pairs would suppress many informative negatives and weaken contrastive discrimination. Pairs involving negative or uncertain states therefore retain unit weight.

\paragraph{Weighted bidirectional contrastive learning.} Let $z_{ki}$ denote the image-sentence logit obtained from the PLRS-corrected similarity map for sentence $q_k$ and image $x_i$. For compactness, define the weighted exponential term as

\begin{equation}
\phi_{ki}
=
W_{ki}
\exp\left(
\frac{z_{ki}}{\tau}
\right),
\label{eq:phi}
\end{equation}

where $\tau$ denotes the contrastive temperature. When $W_{ki}=1$, the term is identical to its standard contrastive counterpart. When $W_{ki}<1$, only the contribution of the selected negative pair is reduced.

Each sentence has one source image. The weighted text-to-image objective is

\begin{equation}
\mathcal{L}_{\mathrm{T2I}}^{w}
=
-\frac{1}{K}
\sum_{k=1}^{K}
\log
\frac{
\exp\left(
z_{k,g(k)}/\tau
\right)
}{
\exp\left(
z_{k,g(k)}/\tau
\right)
+
\sum_{i\neq g(k)}
\phi_{ki}
}.
\label{eq:t2i}
\end{equation}

The numerator in Eq.~\eqref{eq:t2i} remains the original
source image, while IC-SFNS rescales only selected negative
terms in the denominator.

For image $x_i$, let
$\mathcal P_i=\{k\mid g(k)=i\}$ denote its positive sentence
set.
The weighted image-to-text objective is

\begin{equation}
\mathcal{L}_{\mathrm{I2T}}^{w}
=
-\frac{1}{K}
\sum_{i=1}^{B}
\sum_{k\in\mathcal P_i}
\log
\frac{
\exp\left(
z_{ki}/\tau
\right)
}{
\exp\left(
z_{ki}/\tau
\right)
+
\sum_{m:g(m)\neq i}
\phi_{mi}
}.
\label{eq:i2t}
\end{equation}

In both directions, the original positive targets and
numerators remain unchanged.
IC-SFNS only rescales selected cross-patient negative terms
in the denominators, thereby calibrating contrastive pressure
without redefining pair identities.
It introduces no trainable parameters and is used only during
training. The final dual-calibration objective is

\begin{equation}
\mathcal{L}
=
\mathcal{L}_{\mathrm{T2I}}^{w}
+
\mathcal{L}_{\mathrm{I2T}}^{w}.
\label{eq:objective}
\end{equation}

\section{Experiments}
We evaluate PLRS-IC on six public chest X-ray datasets under nine zero-shot settings covering classification, grounding, and segmentation. These tasks jointly assess whether local similarity calibration and global supervision calibration improve both semantic discrimination and spatial correspondence. Training uses only paired radiographs and reports, without task-specific class labels, bounding boxes, or segmentation masks.

\begin{table*}[!t]
\centering
\fontsize{9.2}{10.5}\selectfont
\setlength{\tabcolsep}{1.8pt}
\renewcommand{\arraystretch}{1.03}
\begin{tabular*}{\textwidth}{
@{\hspace{8pt}}l@{\extracolsep{\fill}}cccccccc@{\hspace{8pt}}}
\toprule
\multirow{2}{*}{Method} &
\multicolumn{6}{c}{Classification} &
\multicolumn{2}{c}{Segmentation} \\
\cmidrule(lr){2-7}\cmidrule(lr){8-9}
& Open-I & ChestXray14 & CheXpert & ChestXDet10
& SIIM & RSNA & SIIM & RSNA \\
\midrule
GLoRIA
& 0.589 & 0.610 & 0.750 & 0.645
& -- & -- & -- & 0.347 \\
BioViL-T
& 0.702 & 0.729 & 0.789 & 0.708
& -- & -- & -- & -- \\
MedKLIP
& 0.759 & 0.726 & 0.879 & 0.713
& 0.897 & \textbf{0.869} & 0.044 & 0.465 \\
KAD
& 0.807 & 0.789 & 0.905 & 0.735
& -- & -- & -- & -- \\
CARZero
& 0.838 & \underline{0.811} & \textbf{0.923} & \textbf{0.796}
& \underline{0.924} & 0.747 & \underline{0.100} & 0.540 \\
RadZero
& \underline{0.846} & 0.807 & 0.903 & 0.785
& 0.916 & 0.842 & 0.092 & \underline{0.562} \\
\textbf{PLRS-IC}
& \textbf{0.854} & \textbf{0.813} & \underline{0.911}
& \underline{0.792} & \textbf{0.929} & \underline{0.854}
& \textbf{0.114} & \textbf{0.573} \\
\bottomrule
\end{tabular*}
\caption{Zero-shot classification AUROC and segmentation Dice scores.
The best and second-best results are shown in bold and underlined,
respectively.}
\label{tab:classification_segmentation}
\end{table*}

\begin{table*}[!t]
\centering
\fontsize{9.2}{10.5}\selectfont
\setlength{\tabcolsep}{1.8pt}
\renewcommand{\arraystretch}{1.03}
\begin{tabular*}{\textwidth}{
@{\hspace{8pt}}l@{\extracolsep{\fill}}ccccccccccc@{\hspace{8pt}}}
\toprule
\multirow{2}{*}{Method} & \multicolumn{11}{c}{Grounding} \\
\cmidrule(lr){2-12}
& Mean & ATE & CALC & CONS & EFF & EMPH
& FIB & FX & MASS & NOD & PTX \\
\midrule
GLoRIA
& 0.367 & 0.479 & 0.053 & 0.737 & 0.528 & 0.667
& 0.366 & 0.013 & 0.533 & 0.156 & 0.143 \\
BioViL-T
& 0.351 & 0.438 & 0.000 & 0.630 & 0.504 & \underline{0.846}
& 0.390 & 0.026 & 0.500 & 0.000 & 0.171 \\
MedKLIP
& 0.481 & 0.625 & 0.132 & \textbf{0.837} & 0.675 & 0.734
& 0.305 & \textbf{0.224} & \underline{0.733} & 0.312 & 0.229 \\
KAD
& 0.391 & \underline{0.646} & 0.132 & 0.699 & 0.618 & 0.644
& 0.244 & \underline{0.199} & 0.267 & \underline{0.316} & 0.143 \\
CARZero
& \underline{0.543} & 0.604 & 0.184 & 0.824 & 0.782
& \underline{0.846} & \textbf{0.561} & 0.184 & 0.700
& 0.286 & \textbf{0.457} \\
RadZero
& 0.535 & 0.604 & \underline{0.237} & 0.806
& \underline{0.794} & \textbf{0.897} & 0.427 & 0.184
& \underline{0.733} & \textbf{0.325} & \underline{0.343} \\
\textbf{PLRS-IC}
& \textbf{0.557} & \textbf{0.667} & \textbf{0.263}
& \underline{0.827} & \textbf{0.806} & 0.821
& \underline{0.463} & 0.171 & \textbf{0.767}
& \textbf{0.325} & \textbf{0.457} \\
\bottomrule
\end{tabular*}
\caption{Zero-shot Pointing Game scores on ChestXDet10.
Mean and per-finding results are reported.
The best and second-best results are shown in bold and underlined,
respectively.}
\label{tab:grounding}
\end{table*}

\subsection{Experimental Setup}

\paragraph{Training Data.}
We train PLRS-IC on the official training split of MIMIC-CXR~\cite{mimiccxr}. The dataset contains 377,110 radiographs from 227,835 studies and 65,379 patients. Each study includes one report and one or more frontal or lateral radiographs. We retain the findings and impression sections and divide each report into finding sentences. Every radiograph is associated with the sentences from its source study.

RadGraph~\cite{radgraph} is applied offline to extract clinical concepts and polarity labels. The concepts are normalized and mapped to a fixed inventory of 25 canonical chest X-ray concepts. The inventory and information-content prior are constructed only from the training split. Available projection labels are retained for PLRS.

\paragraph{Evaluation Protocol.}
We evaluate classification on Open-I~\cite{openi}, ChestXray14~\cite{cxr14}, CheXpert~\cite{chexpert}, ChestXDet10~\cite{cxd10}, SIIM~\cite{siim}, and RSNA~\cite{rsna}. A sigmoid function converts each image-sentence logit into a similarity probability, and performance is measured by AUROC.

Grounding is evaluated on ChestXDet10. For each query, the corrected patch-text similarity map is interpolated to the input resolution. Pointing Game measures whether the maximum response lies inside the annotated bounding box. We report the mean score and the ten finding-specific scores.

Segmentation is evaluated on SIIM and RSNA. The interpolated map is thresholded to produce a binary mask. Dice is calculated on positive samples using the threshold-selection protocol of RadZero~\cite{radzero}.

\paragraph{Implementation Details.}
We use XrayDINOv2~\cite{dinov2} as the image encoder and all-mpnet-base-v2~\cite{mpnet,sentencebert} as the text encoder. All radiographs are resized to $224\times224$ pixels. PLRS uses residual rank $d_r=8$ and fixed residual scale $\gamma=0.10$. IC-SFNS uses $\alpha=2$, and its information-content prior remains fixed throughout training. The model is trained for 20 epochs on two NVIDIA RTX 5090 GPUs with a batch size of 128 per GPU. We optimize the model with AdamW using a learning rate of $1\times10^{-5}$ and bfloat16 precision.

\subsection{Results and Analysis}

We compare PLRS-IC with GLoRIA~\cite{gloria}, BioViL-T~\cite{biovilt}, MedKLIP~\cite{medklip}, KAD~\cite{kad}, CARZero~\cite{carzero}, and RadZero(224px)~\cite{radzero}.

\paragraph{Classification.}
Table~\ref{tab:classification_segmentation} shows that PLRS-IC improves RadZero on all six classification datasets, with gains of 0.006--0.013 and an increase in mean AUROC from 0.850 to 0.859. It ranks first on Open-I, ChestXray14, and SIIM and second on the remaining datasets. The consistent gains across different label spaces and acquisition settings indicate improved cross-dataset zero-shot generalization. Because PLRS-IC is trained only on MIMIC-CXR and evaluated without target-domain fine-tuning, these improvements reflect transferable alignment rather than benchmark-specific adaptation.

\paragraph{Segmentation.}
PLRS-IC achieves the best Dice scores of 0.114 on SIIM and 0.573 on RSNA, exceeding the previous best results by 0.014 and 0.011, respectively. Because Dice evaluates the full predicted region rather than only the peak response, these gains indicate that the corrected similarity maps produce more spatially coherent regions after thresholding. The improvements on both datasets show that the benefit extends beyond peak localization to the spatial extent of the recovered findings. Notably, these results are obtained without task-specific segmentation supervision, suggesting that dual calibration improves the transferability of the learned spatial correspondence.

\paragraph{Grounding.}
Table~\ref{tab:grounding} shows that PLRS-IC achieves the
highest mean Pointing Game score of 0.557, exceeding
RadZero by 0.022 and the previous best result by 0.014.
It improves seven of ten findings.
The largest gains over RadZero occur for pneumothorax
($+0.114$), atelectasis ($+0.063$), fibrosis ($+0.036$),
and mass ($+0.034$), showing that the overall improvement
is not driven by a single category.
Since Pointing Game depends on the location of the strongest
response, these gains indicate that the calibrated similarity
maps place their peak activations more accurately within the
annotated regions.
Nodule remains unchanged, while emphysema and fracture
decline, indicating that the benefit is substantial overall but
not uniform across categories.

\begin{figure}[H]
    \centering
    \includegraphics[width=\columnwidth]{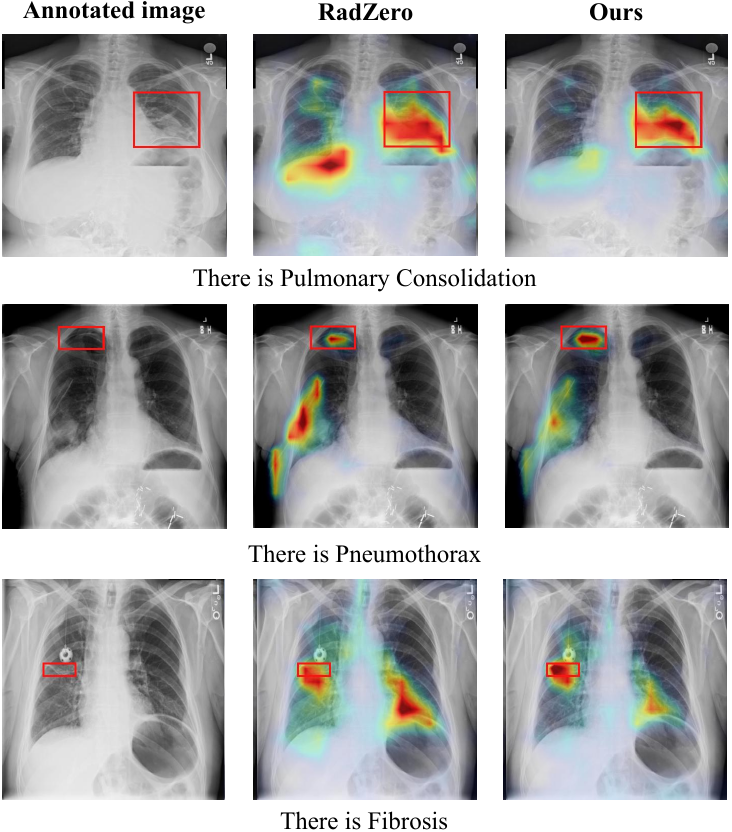}
    \caption{
    \textbf{Qualitative comparison of similarity maps generated by PLRS-IC and RadZero.} Red bounding boxes indicate the ground-truth regions, and heatmaps show query-conditioned responses.
    }
    \label{fig:grounding_visualization}
\end{figure}

\paragraph{Qualitative Grounding Analysis.}
Figure~\ref{fig:grounding_visualization} compares query-conditioned similarity maps from RadZero and PLRS-IC for pulmonary consolidation, pneumothorax, and fibrosis. RadZero often exhibits broad or competing responses outside the annotated regions, indicating that the strongest activations are not aligned with the queried finding. In contrast, PLRS-IC suppresses off-target responses and concentrates activations within or closer to the target boxes. This pattern remains consistent across findings with different visual appearances and spatial extents. These results complement the Pointing Game improvements and suggest that dual calibration produces more spatially selective similarity maps and more precise phrase-region correspondence.

\newpage

\begin{figure}[!t]
    \centering
    \includegraphics[width=\columnwidth]{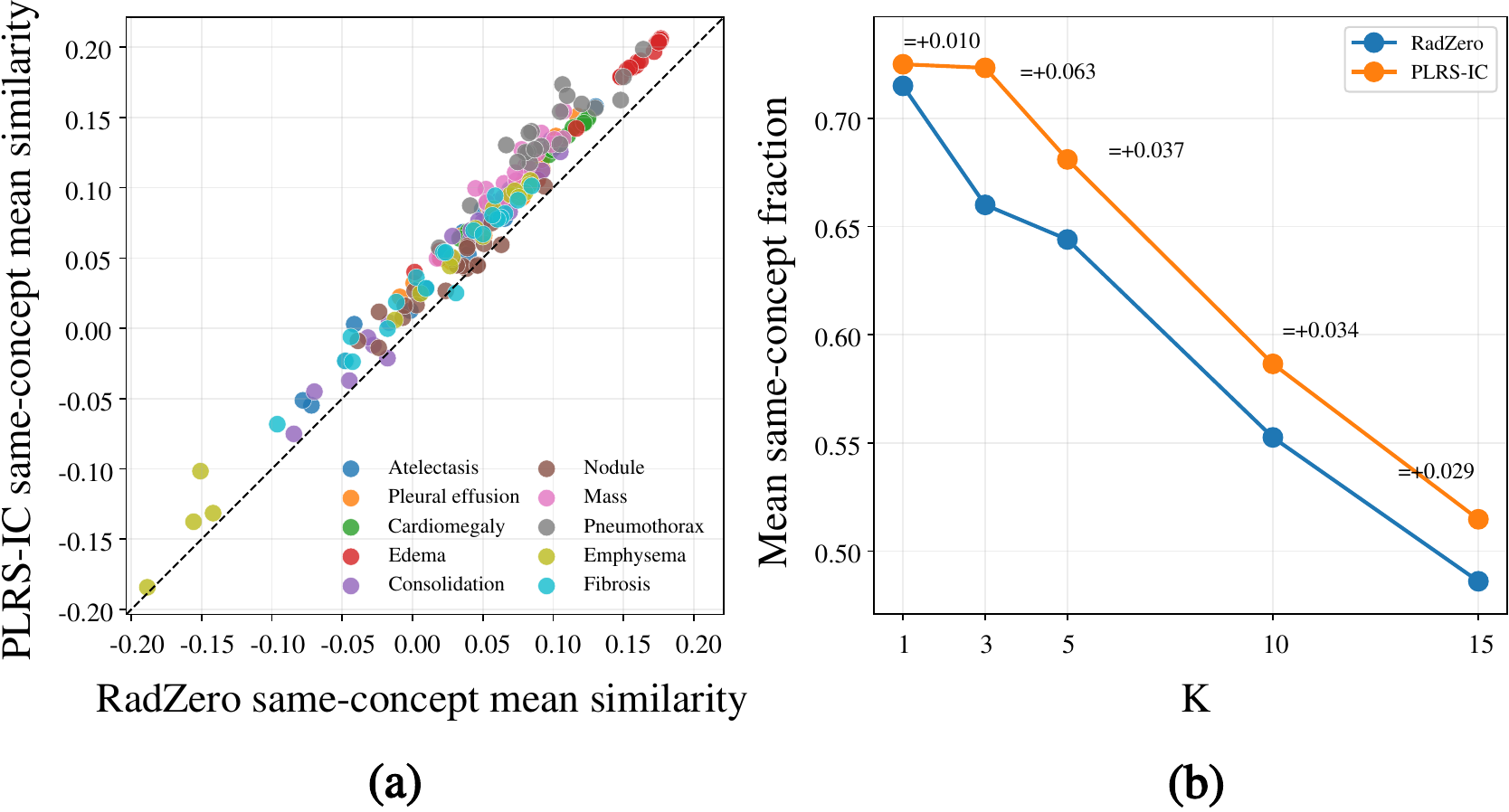}
    \caption{
    \textbf{Cross-patient semantic relation analysis.}
    (a) Mean similarity to same-concept cross-patient images.
    (b) Same-concept fraction among top-$K$ retrieved images.
    Images from the anchor patient are excluded.
    }
    \label{fig:cross_patient_analysis}
\end{figure}

\paragraph{Cross-Patient Semantic Analysis.}
Figure~\ref{fig:cross_patient_analysis} provides two
complementary analyses using 200 anchors from ten concepts
and identical cross-patient candidate sets for both models.
In Figure~\ref{fig:cross_patient_analysis}(a), each point
compares the mean similarity of one anchor to same-concept
cross-patient images under RadZero and PLRS-IC.
Points above the diagonal indicate higher similarity under
PLRS-IC, whereas points below it indicate lower similarity.
PLRS-IC places 195 of 200 anchors above the diagonal and
increases the mean similarity by 0.0267.
Figure~\ref{fig:cross_patient_analysis}(b) reports the mean
fraction of same-concept images among the top-$K$ retrieved
cross-patient candidates, where $K$ denotes the number of
highest-ranked images considered.
PLRS-IC outperforms RadZero at every evaluated
$K=1,3,5,10,$ and $15$, with gains of 0.010, 0.063,
0.037, 0.034, and 0.029, respectively.
These results show that PLRS-IC assigns stronger similarity
to same-concept images and ranks them earlier in
cross-patient retrieval.

\subsection{Ablation Studies}

We evaluate component contributions, projection conditioning, and the IC-SFNS exponent $\alpha$. \emph{Cls. Avg.} denotes the mean AUROC across six classification datasets, and \emph{CXD10} denotes the mean ChestXDet10 Pointing Game score. Non-target settings are fixed within each comparison.

\paragraph{Component Contributions.}
Table~\ref{tab:component_ablation} shows complementary task profiles. PLRS yields the larger grounding gain, consistent with its direct calibration of patch-text similarity maps. IC-SFNS produces stronger segmentation gains, suggesting that calibrating cross-patient contrastive pressure also changes the within-pair spatial correspondence learned through the shared image-sentence logits. This effect is learned during training even though IC-SFNS introduces no inference-time operation. Their combination achieves the best classification, grounding, and SIIM results; on RSNA, it is 0.001 below IC-SFNS alone.

\begin{table}[!t]
\centering
\fontsize{8.4}{9.2}\selectfont
\setlength{\tabcolsep}{2.0pt}
\renewcommand{\arraystretch}{1.06}
\begin{tabular*}{\columnwidth}{
@{\hspace{6pt}}@{\extracolsep{\fill}}cccccc@{\hspace{6pt}}}
\toprule
\multirow{2}{*}{PLRS} &
\multirow{2}{*}{IC-SFNS} &
\multirow{2}{*}{Cls. Avg.} &
\multirow{2}{*}{CXD10} &
\multicolumn{2}{c}{Seg. Dice} \\
\cmidrule(lr){5-6}
& & & & SIIM & RSNA \\
\midrule
\checkmark &            &
0.855 & 0.552 & 0.105 & 0.565 \\
           & \checkmark &
0.854 & 0.539 & 0.107 & \textbf{0.574} \\
\checkmark & \checkmark &
\textbf{0.859} & \textbf{0.557} &
\textbf{0.114} & 0.573 \\
\bottomrule
\end{tabular*}
\caption{Component ablation of PLRS and IC-SFNS.}
\label{tab:component_ablation}
\end{table}

\begin{table}[!t]
\centering
\fontsize{8.4}{9.2}\selectfont
\setlength{\tabcolsep}{2.0pt}
\renewcommand{\arraystretch}{1.06}
\begin{tabular*}{\columnwidth}{
@{\hspace{6pt}}@{\extracolsep{\fill}}lcccc@{\hspace{6pt}}}
\toprule
\multirow{2}{*}{Variant} &
\multirow{2}{*}{Cls. Avg.} &
\multirow{2}{*}{CXD10} &
\multicolumn{2}{c}{Seg. Dice} \\
\cmidrule(lr){4-5}
& & & SIIM & RSNA \\
\midrule
Shared
& 0.857 & 0.545 & 0.106 & 0.566 \\
View-specific
& \textbf{0.859} & \textbf{0.557}
& \textbf{0.114} & \textbf{0.573} \\
\bottomrule
\end{tabular*}
\caption{Ablation of projection conditioning.}
\label{tab:projection_ablation}
\end{table}

\begin{table}[!t]
\centering
\fontsize{8.4}{9.2}\selectfont
\setlength{\tabcolsep}{2.0pt}
\renewcommand{\arraystretch}{1.06}
\begin{tabular*}{\columnwidth}{
@{\hspace{6pt}}@{\extracolsep{\fill}}ccccc@{\hspace{6pt}}}
\toprule
\multirow{2}{*}{$\alpha$} &
\multirow{2}{*}{Cls. Avg.} &
\multirow{2}{*}{CXD10} &
\multicolumn{2}{c}{Seg. Dice} \\
\cmidrule(lr){4-5}
& & & SIIM & RSNA \\
\midrule
1 & 0.854 & 0.532 & 0.103 & 0.566 \\
2 & \textbf{0.859} & \textbf{0.557}
  & 0.114 & \textbf{0.573} \\
3 & 0.855 & 0.544 & \textbf{0.121} & 0.567 \\
\bottomrule
\end{tabular*}
\caption{Effect of the IC-SFNS exponent $\alpha$.}
\label{tab:alpha_ablation}
\end{table}

\paragraph{Projection Conditioning.}
Table~\ref{tab:projection_ablation} compares a shared low-rank correction with separate frontal and lateral projections. Relative to the shared variant, view-specific projections improve classification and grounding by 0.002 and 0.012, while SIIM and RSNA Dice increase by 0.008 and 0.007, respectively. The largest gain occurs in grounding, supporting the benefit of explicit projection conditioning over a shared low-rank correction. Both variants activate only one visual projection per image and therefore require the same per-image residual computation.

\paragraph{Information-Content Exponent.}
Table~\ref{tab:alpha_ablation} evaluates the attenuation strength controlled by $\alpha$. Increasing $\alpha$ from 1 to 2 improves every metric, whereas $\alpha=3$ further improves SIIM Dice but degrades classification, grounding, and RSNA Dice. Weak attenuation at $\alpha=1$ leaves excessive unreliable negative pressure, whereas stronger attenuation at $\alpha=3$ may weaken useful distinctions between clinically non-equivalent patients. We therefore use $\alpha=2$, which provides the best overall balance and indicates that unreliable negatives should be calibrated rather than maximally suppressed.

\FloatBarrier

\section{Conclusion}
We present PLRS-IC, a unified dual-calibration framework for two intertwined ambiguities in chest X-ray vision-language alignment, including projection-induced ambiguity in local patch-text matching and cross-patient semantic ambiguity in global contrastive optimization. PLRS applies a bounded, projection-conditioned low-rank residual to calibrate local similarity, while IC-SFNS uses a corpus-derived information-content prior to calibrate selected global negative contributions without changing their original pair assignments. Extensive experiments on nine zero-shot benchmark settings, covering classification, grounding, and segmentation, demonstrate the effectiveness of PLRS-IC across diverse chest X-ray understanding tasks.

\clearpage

\bibliography{references}

\end{document}